# Safety-Oriented Calibration and Evaluation of the Intelligent Driver Model

**Kingsley Adjenughwure**
Department of Sustainable Urban Mobility and Safety
TNO, The Hague, The Netherlands
Email: kingsley.adjenughwure@tno.nl

**Arturo Tejada**
Department of Integrated Vehicle Safety
TNO, Helmond, The Netherlands
Email: arturo.tejadaruiz@tno.nl

**Pedro F. V. Oliveira**
Department of Integrated Vehicle Safety
Company: TNO, Netherlands
Email: pedro.vieiraoliveira@tno.nl

**Jeroen Hogema**
Department of Integrated Vehicle Safety
TNO, Helmond, The Netherlands
Email: jeroen.hogema@tno.nl

**Gerdien Klunder**
Department of Sustainable Urban Mobility and Safety
TNO, The Hague, The Netherlands
Email: gerdien.klunder@tno.nl

**ABSTRACT**
Many car-following models like the Intelligent Driver Model (IDM) incorporate important aspects of safety in their definitions, such as collision-free driving and keeping safe distances, implying that drivers are safety conscious when driving. Despite their safety-oriented nature, when calibrating and evaluating these models, the main objective of most studies is to find model parameters that minimize the error in observed measurements like spacing and speed while studies specifically focused on calibrating and evaluating unobserved safe behavior captured by the parameters of the model are scarce. Most studies on calibration and evaluation of the IDM do not check if the observed driving behavior (i.e. spacing) are within the model estimated unobserved safety thresholds (i.e. desired safety spacing) or what parameters are important for safety. This limits their application for safety driven traffic simulations.

To fill this gap, this paper first proposes a simple metric to evaluate driver compliance with the safety thresholds of the IDM model. Specifically, we evaluate driver compliance to their desired safety spacing, speed and safe time gap. Next, a method to enforce compliance to the safety threshold during model calibration is proposed.

The proposed compliance metric and the calibration approach is tested using Dutch highway trajectory data obtained from a driving simulator experiment and two drones. The results show that compliance to the IDM safety threshold greatly depends on braking capability with a median compliance between 38% and 90% of driving time, indicating that drivers can only partially follow the IDM safety threshold in reality.

**Keywords:** Car-following model**,** Intelligent Driver Model, model calibration, safety evaluation, safe driving behavior

**INTRODUCTION**

When performing safety assessment using traffic simulations, it is often required to define a reference (safe) driving behavior. However, defining this driving behavior is not a trivial task and the fact that the human driver is still considered as the safety benchmark for other driving systems further complicates this task. This task requires defining safe driving from a human perspective so that the safety performance of any new Automated Driving System (ADS) can be directly compared to that of a human *(1)*. There are many on-going research efforts on developing models for various aspects of human driving. One of the most studied aspects of driving is car following (CF), which describes the interactions with the preceding vehicle in the same lane. This is an important first step in driver modelling and serves as input to other aspects of driving such as lane change and gap acceptance. To this end, many CF models such as the Intelligent Driver Model (IDM), Gipps model, Wiedemann and Full velocity difference model (FVDM) have been developed over the years *(2,3)*. The parameters of these models have also been calibrated using data from naturalistic driving studies, simulator experiments and drones *(3, 4, 5)*. Most of these models incorporate important aspects of safe driving behavior in their definitions, such as collision-free driving, and drivers trying to keep safe distances *(6)*. However, when calibrating and evaluating these models, the main objective of most studies is to find model parameters that minimize the error in observed measurements like spacing and speed *(6,7)*.

Specifically, efforts have been focused on the model's ability to reproduce general driving behavior (mostly to predict traffic flow, speed, density or full trajectories) while studies aimed at calibrating CF models specifically for understanding safe driving behavior are relatively scarce. Even when safety is the main focus of the study, such as done in *(4)*, safety aspects of the model are still not explicitly included in the calibration process. Most of the safety-driven calibration studies focus on classifying drivers into aggressive and cautious drivers using the already calibrated parameters *(8)*, while others use simple surrogate safety metrics (SSMs) such as time gap (TG) and/or time to collision (TTC) to incorporate driver safety without using the driver model parameters *(5)*.

In our opinion, neither of those approaches for incorporating safety during calibration or evaluation is sufficient for safety driven traffic simulation. The reason is that two drivers (one aggressive and the other cautious) with the same TTC value can have different collision probabilities depending on their unobserved driving capability (i.e. how fast and how hard they are able to brake). This is also the reason why CF models which are designed to be collision-free follow the principle that safety is not only dependent on observed physical measurements such as spacing or only on drivers' (aggressiveness) parameters such as desired speed. Such models make use of an unobserved safety metric which combines drivers' unobserved behavior and capabilities (i.e. their parameters) with the observed physical measurements (i.e. speed and spacing). This unobserved safety metric is actually what is embedded in most CF models as a 'collision avoidance' mechanism. For example, the IDM uses the desired safety spacing as the unobserved safety metric to compute acceleration implying that drivers are inherently safety-conscious when driving *(6)*.

Despite the importance of this unobserved safety metric for the IDM, most studies on calibration and evaluation of the model do not check if the observed driving behavior (i.e. spacing and speed) are within the model estimated unobserved safety thresholds or what parameters are important for maintaining the threshold. This limits the applications of such models for safety driven traffic simulations *(4)*.

To fill the gap in IDM model calibration and evaluation specifically for safety, this paper first proposes a simple metric to evaluate driver compliance with the safety thresholds of the IDM model (as defined by its calibrated parameters). Specifically, we evaluate driver compliance to their estimated desired safety spacing and safe time gap (assuming that the model parameters are the best for representing the driver). Next, we propose a method to incorporate compliance to the safety threshold in the model calibration process such that the level of compliance to the safety threshold can be increased (or reduced).

To the best of our knowledge, this research is the first to evaluate drivers compliance to the safety thresholds imposed by IDM parameters. Secondly, this research is the first to incorporate the compliance to the safety threshold in the IDM model calibration process thus contributing towards the evaluation and calibration of the IDM specifically for use in safety driven traffic simulations.

The paper is organized as follows. In the first section, the state of the art in safety calibration and evaluation of the IDM model is presented. In the second section, the IDM and its variant IDM plus (IDM+) are presented. After that, the proposed safety focused objective function and evaluation metric are described. Next, both IDM models are calibrated and evaluated with the proposed safety objective function and safety threshold compliance metric. The paper concludes with a discussion on IDM model calibration and evaluation for safety and offers future research directions and applications in traffic simulations.

**STATE OF THE ART IN TRAJECTORY BASED IDM CALIBRATION AND EVALUATION FOR SAFETY**

Kesting et al. calibrated the IDM model using data from instrumented vehicles (*3*). They used three objective functions based on spacing error. They compared the performance of the IDM model with the velocity difference model in terms of their ability to replicate trajectory (minimize error in spacing) and also to avoid collision. However, they did not explicitly consider driver behavior compliance to the model's safety threshold during their calibration and evaluation, but rather included a large penalty for collision because the velocity difference model was not collision free.

Similarly, Punzo and Simonelli (*5*) and Punzo et al (*9*) have extensively calibrated the IDM using various objective functions. In their work, they compared calibrations made based on spacing versus those of speed and time gap. They concluded that spacing gives lower calibration errors compared to speed. Although the authors also used the time gap as a safety metric for calibration, driver behavior compliance to the model's safety threshold was not evaluated. Their main focus was not safety but to show that spacing is better for calibrating CF models.

Berghaus et al (*4*) calibrated the IDM based on driving simulator data. They studied the influence of driver characteristics such as age and gender on driving behavior parameters. This work also looked at some safety aspects of model calibration by examining the applicability of regular CF models in extreme traffic situations such as hard-braking. The study had a safety focus but the calibration objective was based on minimizing the error in speed, spacing and acceleration. Furthermore, the study also classified drivers as aggressive or careful based on estimated parameters. However, there was no evaluation or discussion of driver compliance to the model safety threshold defined by the model parameters.

The idea to consider safety in the CF calibration process has also been recently applied in Dai et al. (*10*) for calibrating Adaptive Cruise Control (ACC) models. In their work, various SSMs such as Time Exposed Time to collision (TET) and Rear end Collision Risk Index (RCRI) were included in the control objective of the ACC. The results were then compared to the basic ACC and IDM model in terms of minimizing risk during artificial perturbation in a platoon. They concluded that explicitly considering SSMs in ACC calibration reduces rear-end collision risk compared to the basic calibration. However, this study focused on calibrating ACC behavior and not the human (IDM) behavior.

Similarly, Liu et al. (*11*) have explicitly considered safety in the calibration of ACC parameters. In their study, the ACC model parameters were calibrated to minimize the safety area below a certain time to collision (TTC) threshold and the difference between the distance gap and the minimum distance gap as calculated by the Responsibility-Sensitive Safety model (RSS) model. The results showed that this approach improved the safety performance of the ACC model during cut-in scenarios compared to the original ACC model. Although this work above includes safety in the calibration and evaluation, they focused on making ACC much safer and not on using the IDM to understand human safe driving behavior.

For modeling safe human driving behavior, our literature review only found one research that explicitly computed the desired spacing of the IDM and compared it to actual spacing (*12*). Again this research only focused on verifying if the speed-spacing relationship produced by CF models such as the IDM are closer to the speed-spacing relationship in reality. Safety was not the focus of the calibration but traffic efficiency. Also, the evaluation metric proposed does not measure the level of compliance to the IDM safety threshold, but rather measures the error in the speed-spacing relationship.

**IDM MODELS**

There are several variants of the IDM model in literature, however, we focus only on the original model (*6*) and a variant of the IDM+ (*13*) that was specifically proposed to improve estimation of traffic capacity (i.e. efficiency driven). This allows us to compare a safety driven model formulation with an efficiency driven model formulation. The model formulations are discussed below.

**IDM**

The IDM is one of the most popular car-following models used for modelling car-following behavior. The model describes the acceleration behavior of a single vehicle (follower) on a single lane when driving alone or when reacting to a lead vehicle (leader). The model assumes that the follower's acceleration depends primarily on its velocity, the relative speed to the lead vehicle and the net distance gap between the vehicles. The driving principle of the model is such that the "intelligent driver" will accelerate to reach its desired velocity when driving freely (free flow component) and it will decelerate when approaching too fast or driving too close to its leader (car-following component).

The model is defined in its generic form for a vehicle $\alpha$ as (*6*):

$$\dot{v}_\alpha = a^{(\alpha)} \left[ 1 - \left( \frac{v_\alpha}{v_0^{(\alpha)}} \right)^\delta - \left( \frac{s^*(v_\alpha, \Delta v_\alpha)}{s_\alpha} \right)^2 \right] \qquad (1)$$

$$s^*(v, \Delta v) = s_0^{(\alpha)} + s_1^{(\alpha)} \sqrt{\frac{v}{v_0^{(\alpha)}}} + T^\alpha v + \frac{v \Delta v}{2\sqrt{a^{(\alpha)} b^{(\alpha)}}} \qquad (2)$$

- $a^{(\alpha)}$ is the maximum acceleration
- $b^{(\alpha)}$ is the desired deceleration
- $v_\alpha$ is the current velocity
- $v_0^{(\alpha)}$ is the desired velocity
- $\delta$ is the acceleration exponent
- $\Delta v_\alpha$ is the approaching rate
- $s_\alpha$ the net distance gap
- $s^*$ is the desired dynamic distance gap
- $T^\alpha$ is the safe-time headway
- $s_0^{(\alpha)}$ and $s_1^{(\alpha)}$ are the jam distances.

A simplified version of the model is achieved by setting $s_1^{(\alpha)} = 0$ and $\delta = 4$. The model is collision free and is able to replicate realistic car-following behavior on motorways with the emerging complex traffic states and phenomena such as congestion, shock-waves and hysteresis (*3, 6*).

**IDM+**

Authors in (*13*) found that the IDM model proposed in (*6*) gives small capacity values (just below 1900 veh/h/lane) for reasonable values of safe time gaps. Therefore, a simple modification was made to the original IDM model to increase the capacity value to 2200 veh/h/lane. In the modified IDM model, the free-flow and interaction components of **Equation 1** are separated by replacing the addition with the minimum operator. This modified IDM model is called the IDM+ and is defined as:

$$\dot{v}_\alpha = a \cdot min \left[ 1 - \left( \frac{v}{v_0} \right)^\delta, 1 - \left( \frac{s^*(v, \Delta v)}{s} \right)^2 \right] \qquad (3)$$

$$s^*(v, \Delta v) = s^*(v, \Delta v) = s_0 + vT + \frac{v \Delta v}{2\sqrt{ab}} \qquad (4)$$

Where parameters have the same definition as IDM above.

**CURRENT CALIBRATION OBJECTIVES AND MEASURES OF PERFORMANCE**

First we introduce the common calibration objectives found in literature *(5, 9)*. Note that the actual definition of the objective function is not the main focus of the paper. For example, various studies have defined different objective functions based on Measures of Performance (MoP) like spacing, speed and acceleration. The most commonly used objective functions are the Root Mean Square Error (RMS), Normalized Root-Mean-Square-Error (NRMSE), and the Theil's inequality coefficient *(3, 5, 9)*. For simplicity, we choose the NRMSE as it is a commonly used objective function and easily comparable across models. The objective function is the NRMSE of spacing defined as *(9)*:

$$NRMSE\ (s) = \frac{\sqrt{\frac{1}{N} \cdot \sum_{t=1}^{N} [s^{obs}(t) - s^{sim}(t)]^2}}{\sqrt{\frac{1}{N} \cdot \sum_{t=1}^{N} [s^{obs}(t)]^2}} \quad (5)$$

Where $N$ is the number of time-steps in the trajectory, $s^{obs}$ and $s^{sim}$ are the observed and simulated spacing values. The definition of the NRMSE objective function in **Equation 5** is also used for the speed (v) and time-gap (TG) MoPs *(9)* with only the modification of the measured value (i.e., by replacing spacing by v or TG, respectively).

**IDM SAFETY MEASURE OF PERFORMANCE AND CALIBRATION OBJECTIVE**

To explicitly take into account safety in the calibration of the IDM, we propose a simple objective function which uses the dynamic desired safety gap from the IDM as a MoP. The objective function seeks to minimize the error between the desired dynamic spacing of the IDM, $Sstar_i^{req}$ and the simulated dynamic spacing $Sstar_i^{sim}$. The required dynamic spacing is defined as the dynamic spacing desired by the driver assuming that the calibrated parameters are in fact the actual representation of the driver. The advantage of this MoP is that it allows the optimization algorithm to find the optimal (in terms of safety) set of parameters that drivers are required to have to ensure that their observed driving behavior is within the IDM safety thresholds.

The proposed safety objective in terms of the NRMSE is defined as:

$$NRMSE\ (sstar) = \frac{\sqrt{\frac{1}{N} \cdot \sum_{t=1}^{N} [sstar^{req}(t) - sstar^{sim}(t)]^2}}{\sqrt{\frac{1}{N} \cdot \sum_{t=1}^{N} [sstar^{req}(t)]^2}} \quad (6)$$

$$Sstar^{req}(v_{obs}, \Delta v_{obs}) = s_0 + v_{obs}T + \frac{v_{obs}\Delta v_{obs}}{2\sqrt{ab}} \quad (7)$$

$$Sstar^{sim}(v_{sim}, \Delta v_{sim}) = s_0 + v_{sim}T + \frac{v_{sim}\Delta v_{sim}}{2\sqrt{ab}} \quad (8)$$

Since the MoP uses the IDM safety spacing, minimizing the error in the MoP ensures that those parameters are well calibrated towards IDM safe driving. To ensure there is a balance between current efficiency driven MoPs and the proposed safety MoP, we propose to simultaneously minimize both the error in the actual measured spacing (as it is currently done) and the error in the required dynamic safety spacing.

The proposed combined (safety and efficiency) objective functions is defined below for spacing:

$$NRMSE\ (s, sstar) = \alpha \cdot NRMSE\ (s) + \beta \cdot NRMSE\ (sstar) \tag{9}$$

Where $\alpha, \beta$ are the weights assigned to the observed spacing $s$ and the unobserved safety spacing $sstar$.

**IDM SAFETY COMPLIANCE EVALUATION METRIC**

Apart from calibrating the IDM for safety, it is also necessary to evaluate the calibrated model in terms of driver compliance to safety thresholds. There are two reasons to do this. First to provide a way to compare two calibrated models in terms of safety compliance. In fact, we use the safety compliance metric to show that two models can have good performance in spacing error but different performance in safety compliance. Secondly, we use the safety compliance metric to show that a model whose parameters are calibrated for safety using both the proposed safety MoP and spacing MoP will be more safety compliant compared to a model calibrated only for efficiency using the spacing MoP.

To define the safety compliance in the IDM, we first define the required distance gap $s_t^{\text{req}}$ as:

$$s_t^{\text{req}} = SStar_t^{\text{req}} \tag{10}$$

Where $SStar_t^{\text{req}}$ is the model required desired safety spacing at time step $t$ as calculated by the parameters of the model (**Equation 7**). This is the required minimum distance gap to maintain the driver's desired safety threshold. We then compare this safety gap to the actual distance gap which drivers keep. For simplicity, we assume that safer drivers will try to drive close to the required distance gap in order to maintain this safety threshold. Driving with a distance gap smaller than $s_t^{\text{req}}$ will then be considered violating the safety threshold of the model while a larger distance gap than $s_t^{\text{req}}$ will imply that drivers are maintaining the safety threshold imposed by the model. In addition to the dynamic safety, the observed time gap $TG_t^{\text{obs}}$ should be greater than the model estimated safe time gap, $T$ at all times. Finally, we assume that safe drivers will not exceed their desired speed because they consider it to be both the fastest and safest speed they can handle. Therefore, the model's compliance to the IDM safety threshold is defined as:

$$SC(t) = \begin{cases} 1, if\ s_t^{\text{obs}} \geq s_t^{\text{req}}\ and\ TG_t^{\text{obs}} \geq T\ and\ v_t^{\text{obs}} \leq v_0 \\ 0, otherwise \end{cases} \tag{11}$$

Where SC(t) is the safety compliance at time step $t$, $v_t^{\text{obs}}$, $s_t^{\text{obs}}$ and $TG_t^{\text{obs}}$ are the observed speed, distance and time gap, $v_0$ is the estimated desired speed, and $T$ is the estimated safe time gap.

To compare the compliance between trajectories, we use the average compliance per trajectory defined as:

$$\frac{1}{N} \cdot \sum_{t=1}^{N} SC(t) \tag{12}$$

Note that the definition of the safety compliance (**Equation 11**) given above is not the only way to measure the compliance to the safety threshold, but we give this definition as a first attempt to evaluate this property in calibrated models. For example, the compliance can be relaxed by omitting the speed requirement or safe time gap requirement. Also apart from having compliance value in the set {0,1} other definitions are possible for example by using fuzzy logic as proposed in (*14)* where compliance can be in the interval [0,1].

**APPLICATION**

To show the advantage of explicitly calibrating and evaluating IDM for safety, the proposed objective function and safety compliance metric were applied to trajectories of CF events collected by drones in a highway in the Netherlands *(15)* and also extracted from a driving simulator experiment *(16)*.

**Data Collection by Drones**
The raw data consist of recorded video by two drones flown adjacent to each other over the A2 highway near Best in the Netherlands on September 13, 2022, from 1pm till 4pm (off-peak period). The combined field of view of both drones was about 650 m in both directions of traffic. The video was recorded at 25Hz, which is sufficient for estimation of speed and acceleration. The speed limit for this stretch of three-lane highway was 100 km/hr. Since the focus of this paper is on the calibration process, a full description of the data collection and processing is not given here, instead we focus on the data selection, objective function and the model performance. A more detailed description of the data can be found in (*15*).

**Data Collection Driving Simulator Experiment**
A driving simulator study was conducted with 35 participants driving in the TNO driving simulator (*16*). They made trips of about 30 minutes on a simulated Dutch highway in various scenarios, including free driving, CF and responding to a braking lead vehicle.

**Data Selection Drone**
The vehicle trajectories were extracted using third party proprietary software. After this, we selected CF events based on the measured time gap (to ensure that there is a lead vehicle) and duration of the CF event (to ensure enough data for calibration). Therefore, the selected CF events had a time gap between 0.25 s and 3 s and the duration of the CF event was at least 20 s. This resulted in a total of 3395 CF events.

**Data Selection Driving Simulator Experiment**
From the driving simulator data (10 Hz), all CF events were selected, using a criterion of having a time gap below 3 s for a duration of the at least 20 s. This resulted in a total of 136 events.

**Calibration Objective Functions**
The IDM models were calibrated to minimize the classic spacing objective (**Equation 5**) and the proposed objective of minimizing the spacing in combination with the dynamic safety spacing of the IDM (**Equation 9**). For simplicity and ease of comparison, we set $\alpha = 1$ and $\beta=1$ indicating that both safety and efficiency are equally important. The spacing MoP is chosen as benchmark for comparison because it is the recommended MoP found in literature (*3, 5,9*).

**Model Fitting and simulation**
In general two types of model fitting are used in literature. A local fitting based on a trajectory calculated using ground truth data of the follower as input to the model every time step and a global fitting based on a trajectory calculated using ground truth data of the follower as input to the model only in the first time step (*3, 5*). This study uses the global fitting technique as it has been shown to perform better than local fitting in benchmark data sets (*2,3,7*). Therefore, global fitting is used to estimate the complete trajectory of the follower (starting from an initial position and speed). The simulation step is kept the same as the time resolution of the data: 25Hz for the drone data and 10Hz for the driving simulator experiment. This is considered sufficient for safety evaluations (authors in (*7*) have shown that as long as the time interval is less than 1s, the calibration is not significantly affected by the sampling frequency).

**Optimization Algorithm**
The optimization algorithm is an important aspect of the calibration as it determines the quality of the calibrated parameters and speed of the calibration. There are several optimization algorithms already applied in calibration of car-following models. The most popular optimization algorithms used for global fitting of car-following models are the Genetic Algorithm (GA) used in (*3*), (*5*) and (*9*) and the Simultaneous Perturbation Stochastic Approximation (SPSA) algorithm used in (*17*) for comparison with GA. Although these algorithms are widely used, they are not guaranteed to give an optimal solution because they easily get stuck in local minima. To alleviate this problem, authors in (*17*) proposed a global optimization algorithm called DIRECT-SQP. This algorithm combines a direct search for the global optimum with a local search for the local optimum. In this study, the DIRECT-SQP optimization algorithm was used as it has been shown to achieve fast convergence to global optimum in the calibration of various car-following models including the IDM (*17*). Additionally, using a global optimization allows consistency in results and avoids comparing average performance when comparing models. Note that we do not make a comparison of various optimization algorithms as it is not the goal of this paper to compare algorithms but rather to show how to calibrate and evaluate the IDM for safety using any optimization algorithm of choice. A detailed description and performance evaluations of various optimization algorithms can be found in (*17*). The parameters of the optimization algorithm are given in the **Table 1** below.

**Table 1:** Parameters of optimisation algorithm

| Parameters DIRECT-SQP | Value |
|---|---|
| Maximum number of iterations of global searcher (DIRECT) | 50 |
| Maximum function evaluations (DIRECT) | 10, 000 |
| Maximum number of hyper-rectangle divisions (DIRECT) | 10, 000 |
| Minimum possible optimal (PO) hyper-rectangle size or critical size (DIRECT) | 0.01 |
| ConstraintTolerance (SQP) | 1e-10 |
| OptimalityTolerance (SQP) | 1e-10 |
| MaxIterations (SQP) | 2000 |
| Software (DIRECT+SQP) | Matlab 2023a |
| Optimisation Function (SQP) | fmincon |

**Model Parameters**
The data from the simulator experiment contain complete trajectories according to the definition in (*18*), therefore all 6 parameters of the IDM and IDM+ model were calibrated. The drone data was collected during uncongested traffic flow, therefore only 4 parameters were calibrated. The acceleration exponent $\delta$ was fixed to 4 as standard practice (*3,6*) and the stopping distance was set to 2m as the minimum safety distance *(1)*. The parameters and their bounds used for both calibrations are shown in the **Table 2** below:

**Table 2:** Bounds of Parameters of the IDM and IDM+ model

| Parameter | Simulator Experiment Calibration | Drone Experiment Calibration |
|---|---|---|
| Maximum acceleration, $a$ | [0.1, 6.0] | [0.1, 6.0] |
| desired deceleration, $b$ | [0.1, 6.0] | [0.1, 6.0] |
| desired velocity, $v_0$ | [20.0, 40.0] | [20.0, 40.0] |
| acceleration exponent, $\delta$ | [2.0, 4.0] | 4.0 (fixed) |
| jam distance, $s_0$ | [2.0, 5.0] | 2.0 (fixed) |
| jam distance, $s_1$ | 0 (fixed as standard practice) | 0 (fixed as standard practice) |
| safe-time headway, $T$ | [0.5, 6.0] | [0.5, 6.0] |
| Total Number of calibrated parameters | 6 | 4 |

## RESULTS
### Effect of Safety Objective on Model Calibration Errors

We compare the calibration errors of the spacing objective and the proposed combination of spacing and dynamic safety spacing using box-plots. The results in **Figure** 1 show that the spacing errors are within the expected range of 0-30% (*3,7,9*) implying that the optimization algorithm was able to find the right parameters that match the data for both objectives. In terms of magnitude, the spacing errors are slightly smaller when using the proposed safety objective function. This holds regardless of model or data type. This suggests that drivers actually take into account the dynamic safety spacing when following their leader. If this was not the case, the addition of safety would have resulted in very large errors. This is in agreement with the assumption made by the IDM model that drivers actually use the dynamic safety in making decision.

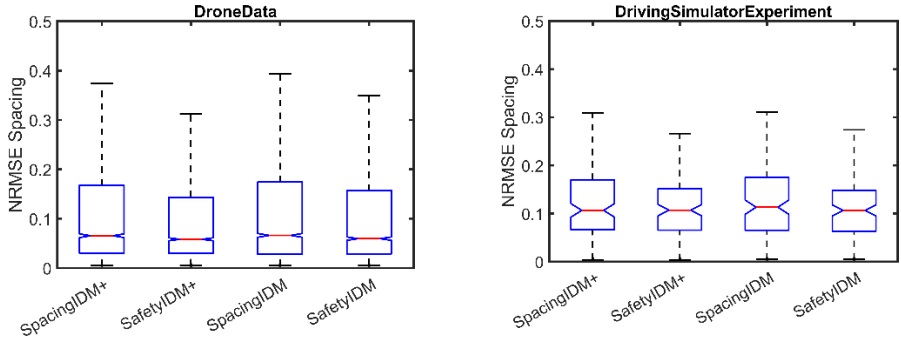

**Figure 1: NRMSE of spacing**

Similarly, the speed error **Figure 2** are as expected lower than the spacing error irrespective of calibration objective, data or model. This is consistent with results from other studies (*3,5,6*). The order of magnitude of the errors are also the similar with all errors less than 10% for the drone data and less than 15% for the driving simulator data. Finally, looking at the time gap errors (**Figure 3**), the proposed safety objective function performs slightly better than the spacing objective. This is mainly due to the better prediction of spacing. Any worse speed prediction is compensated for by a better spacing prediction leading to an overall better prediction of the time gap. Overall, the results reveal that including safety in the calibration does not significantly affect the errors made by the calibrated model both in speed, spacing and time-gaps.

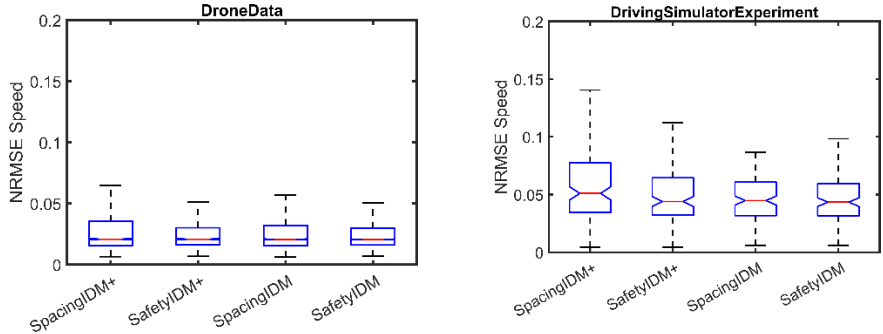

**Figure 2**: **NRMSE of speed**

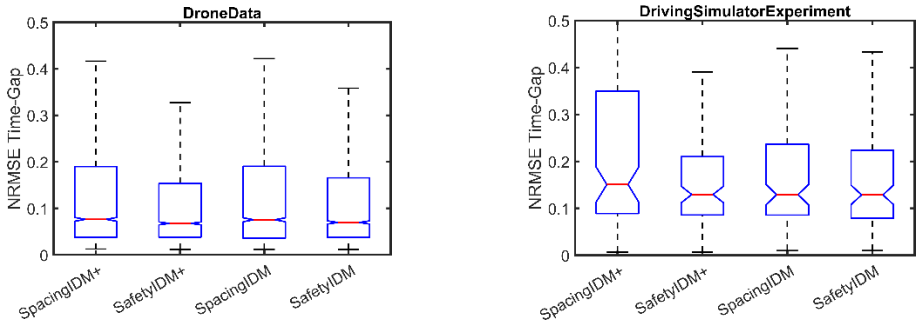

**Figure 3: NRMSE of time gap**

**Effect of Safety Objective on Driver Compliance to IDM Safety Thresholds**
The box plots (**Figure 4**) of the average safety compliance level show clearly the importance of calibrating and evaluating the IDM specifically for safety. The figure reveals many interesting findings. The first finding is that two models may actually have comparable errors in spacing and speed but their safety compliance level maybe entirely different. For example, the proposed safety compliance metric clearly reveals that drivers using the IDM+ are less safety compliant (median 38% in drone data and 50% in simulator data ) compared to drivers using the original IDM model (median 58% and 65%). This finding holds regardless of data or calibration objective. This sharp difference in safety compliance level between the IDM+ and the IDM cannot be directly inferred from the errors in speed, spacing or time-gap because the errors do not say much about safety. This confirms our argument that there is need to evaluate a calibrated model specifically for safety in order to fully understand the safety aspects of the driving behavior represented by the model and its parameters. This finding can be directly applied in a safety simulations, by using the IDM to simulate safer drivers and the IDM+ for less safe drivers.

The second finding is that calibrating for the spacing alone (as is current practice) already shows some level of safety compliance. The results reveal that the median driver compliance to the IDM safety threshold is 65% for the driving simulator experiment data and 55% for the drone data indicating that drivers partially keep the safety thresholds of IDM parameters. This is not surprising because according to the IDM, the gap drivers actually keep is dependent on the level of safety that they are subconsciously trying to keep (i.e. the desired safety spacing). This shows that the drivers may actually be regulating their spacing to meet the IDM safety threshold. On the flip side, the results also show that drivers do not always follow the IDM safety specification which is also not surprising (as the IDM maybe too safe compared to a human driver). This partial compliance safety behavior cannot be inferred from evaluating only errors on physical measurements alone. This also supports our argument for a safety oriented evaluation of calibrated models. In terms of application in safety simulations, the partial compliance behavior can be directly applied by specifying that drivers strictly follow the IDM safety specification only for a proportion of their driving time and can violate it at other times during the simulation.

The third finding is that calibrating specifically for safety increases the safety compliance level for the IDM. This is true for the original IDM both in the drone data and the driving simulator data. The median safety compliance level jumps from (55% for drone data and 65% for simulator data) to around 90% (for both) when using the proposed safety objective. This means that the safety objective function forces the optimization algorithm to find the parameters that make the observed drivers' behavior strictly compliant with the IDM safety threshold (i.e. much safer) something that was not possible using only the spacing objective. These safe parameters can be used in safety simulations for simulating safe drivers or for simulating autonomous vehicles that keep same gap and speed as humans but have different safety parameters. These findings show that calibrating and evaluating a model for safety has lots of potential in better understanding of safe driving behavior which can be directly applied in safety-driven traffic simulations.

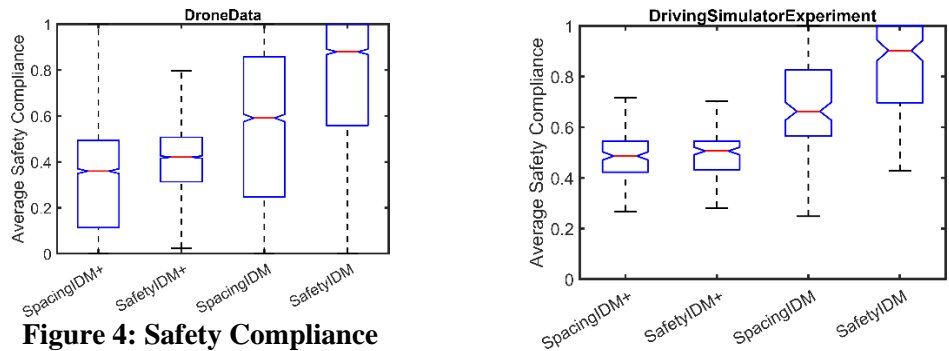

**Figure 4: Safety Compliance level**

**Effect of Safety Objective on Model Calibrated Parameters**

The box plots of the most important parameters of the model, reveal that the parameters of the proposed safety objective function significantly differs from the spacing objective mainly in the maximum acceleration (**Figure 5**) and the comfortable deceleration parameters (**Figure 6**). The median of the maximum acceleration and comfortable deceleration in the safety objective is consistently higher than the spacing objective. This finding holds for both the drone and simulator dataset. This suggests that the current gaps which driver keep are only guaranteed safe (according to the IDM) if drivers are able to brake hard (deceleration parameter) and fast (acceleration parameter). Interestingly, the optimal value for the comfortable deceleration when using the safety objective is the upper-bound for that parameter which is 6 m/s$^2$ (except for the IDM+ which had a slight variation in the value). This immediately raises the question if drivers are comfortable braking at 6 m/s$^2$ which is closer to an emergency braking instead of a comfortable braking *(2,4)*. In our opinion, this high value of comfortable deceleration is not entirely surprising. Our explanation for this high value is that in reality drivers may not be comfortable braking that hard but they still keep unsafe distances and speeds thinking they are safe (i.e. overestimating their capabilities) or they know their capabilities but they accept some level of risk anyways. Both scenarios are entirely plausible and will most likely lead to an emergency braking instead of a comfortable braking during safety critical situations (i.e. hard brake of the leader). This is closer to what happens in reality, suggesting that drivers actually accept some level of risk or overestimate their capabilities.

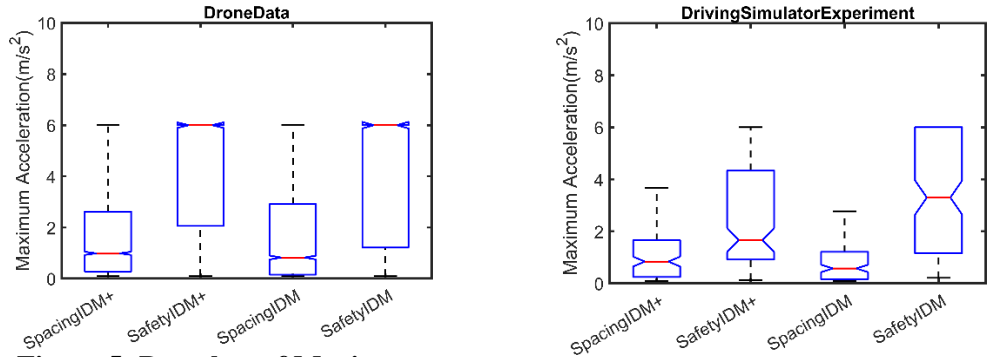

**Figure 5: Box plots of Maximum Acceleration**

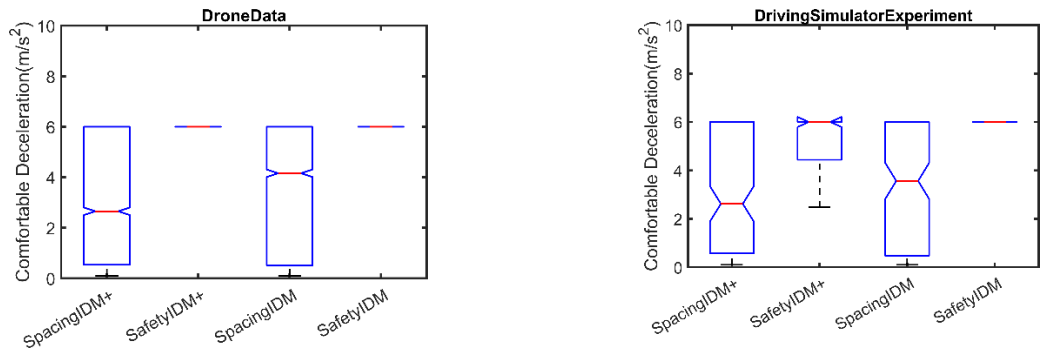

**Figure 6: Box plots of Comfortable Deceleration**

Looking at the median values of the other two parameters we see that the safety objective leads to slightly higher median safe time gap (**Figure 7**), and lower desired speed (**Figure 8**). This is in line with expectation as safe drivers will prefer larger gaps and smaller desired speeds. The results also show that these parameters are not the most important determinant of safety for the IDM as the calibrated values for both objectives have the same order of magnitude and are within expected range for human drivers *(2,4,9)*. This also confirms that the safety objective is able to find safer parameters but also able to detect which parameters are the most relevant for safety.

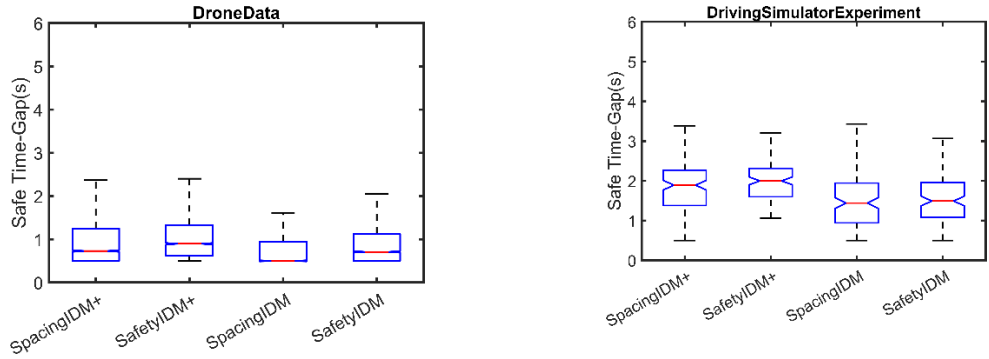
**Figure 7: Box plot of safe time gap**

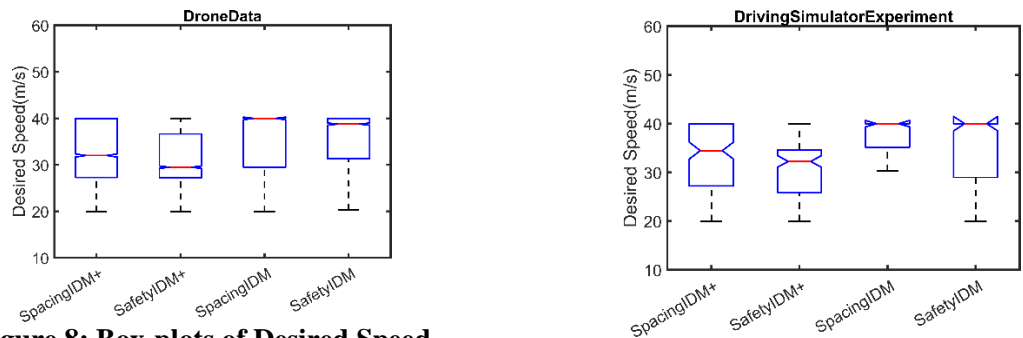
**Figure 8: Box-plots of Desired Speed**

**CONCLUSION**

The IDM is one of the most popular CF models used for modelling car-following behavior on motorways. The model has been extensively studied and its parameters have been calibrated for various purposes such as reproducing general traffic flow dynamics or reproducing trajectories of individual vehicles (drivers). The model possesses many aspects of safety in its definition and parameters such as collision free driving, keeping a safe distance. However, many of the model calibration and evaluation efforts do not specifically focus on how to use the model and its parameters to understand and evaluate human safe driving behavior for use in safety driven traffic simulations. To this end, this paper proposed a simple objective function for the calibration of IDM specifically for safety which minimizes the error in the actual spacing and the dynamic safety spacing derived from the IDM model. Furthermore, a metric to evaluate driver compliance to the IDM safety thresholds was proposed. The objective function and evaluation metric were tested on both drone and driving simulator highway trajectory data using two variants of the IDM.

Our results show that drivers partially comply to the IDM safety threshold (about 65% of their driving time) and this compliance is strongly dependent on braking capability which is controlled by the maximum acceleration and comfortable braking parameters of the IDM. In fact better braking capability ($6m/s^2$) increased their median compliance level to about 90% of driving time. However, this high safety compliance is most likely not possible in reality because most drivers are not comfortable braking that hard except in emergencies. This leads us to conclude that drivers are overestimating their abilities or accepting some level of risk while driving. Our findings have direct application in safety driven traffic simulations such as making drivers safety conscious in only a proportion of their simulation time, correctly setting the parameters of safe and unsafe drivers in the simulation. Another safety application of our finding is to use the proposed safety objective function to design ADS that keep similar spacing, speed and time gaps like humans but possess better braking capabilities than human (as expected). In our opinion, this is the best way to design automated driving systems instead of systems that behave totally different from human drivers (e.g. by keeping large gaps).

## LIMITATIONS AND FUTURE RESEARCH

Although our work reveals many important findings regarding safe driving behavior, there are some limitations that need to be addressed in the future.

First, all definitions of safe driving were based on the assumption that the calibrated IDM parameters are the true representation of the driver (i.e. the optimization algorithm was able to find the right parameters). So the computation of required safety gap uses these parameters as if they were the true parameters the drivers used in making their decision along with the true measured gaps and speed. The estimated compliance level thus contains some margin of error which depends on how much the estimated parameters differ from the true parameters. One way to circumvent this limitation is to use different optimization algorithms or different objective functions which include speed and acceleration like the ones specified in (*9*) to find different optimal parameters and then use the average values in the computation of compliance level.

Secondly, the data used for the drone calibration has a minimum duration of 20 s. This is a relatively short period of time for calibrating long term driver behavior. This limitation was due to the data collection by drone which had short CF duration because of short recording times per vehicle. Also, the drone data does not include all driving regimes, but mostly non congested traffic flow, therefore it is not a complete trajectory according to the definition in *(18)*. The driving simulator data partially solved this issue because it contains much longer and complete CF events. However, the duration is still in order of minutes which is still not enough to make a general statement on long term safe driving behavior. Subsequent studies with much longer duration (e.g. hours) of driving under different traffic conditions are needed to further validate our findings.

Finally, our definition of safety compliance is very strict because it only allows compliance value in the set {0,1} .This can be further improved by using fuzzy logic similar to what was proposed in (*14*) where compliance can be in the interval [0,1]. This is relevant because some drivers maybe violating the threshold just by a small margin while others may have a higher violation margin. This can help to further classify drivers in simulation into various safety compliance levels.

Finally, a more macroscopic validation of our findings is needed. Although we describe how to apply our findings on driver compliance in a traffic simulation, we did not perform an actual traffic simulation to check the effect of this behavior on overall traffic flow, traffic stability and other macroscopic traffic patterns which emerge from the behavior. This will be our next research agenda.

## ACKNOWLEDGMENT

We would like to thank our colleagues Frank Evers, Jeroen Manders, Michiel Braat, and Jorrit Goos for their technical support in acquiring and preprocessing the data.

## AUTHOR CONTRIBUTIONS

The authors confirm contribution to the paper as follows: study conception and design: Kingsley Adjenughwure and Arturo Tejada, data collection: all, analysis and interpretation of results: all, draft manuscript preparation: all. All authors reviewed the results and approved the final version of the manuscript.

## FUNDING INFORMATION

This project was funded by Knowledge Investment Program (KIP) of TNO, KIP-Safe and Automated 2024.